\pdfoutput=1

\documentclass[11pt]{article}

\usepackage[final]{coling}

\usepackage{times}
\usepackage{latexsym}
\usepackage{tablefootnote}
\usepackage[T1]{fontenc}
\usepackage[T5]{fontenc}

\usepackage[utf8]{inputenc}

\usepackage{microtype}

\usepackage{inconsolata}

\usepackage{graphicx}
\usepackage{booktabs}
\usepackage{multirow}
\title{BEIR-NL: Zero-shot Information Retrieval Benchmark
for the Dutch Language}

\author{Nikolay Banar\thanks{indicates equal contribution} \And Ehsan Lotfi\footnotemark[1]  \And Walter Daelemans \\ \AND
        CLiPS, University of Antwerp, Belgium \\ \texttt{\{nicolae.banari, ehsan.lotfi, walter.daelemans\}@uantwerpen.be}}

\begin{document}
\maketitle
\begin{abstract}
Zero-shot evaluation of information retrieval (IR) models is often performed using BEIR; a large and heterogeneous benchmark composed of multiple datasets, covering different retrieval tasks across various domains. Although BEIR has become a standard benchmark for the zero-shot setup, its exclusively English content reduces its utility for underrepresented languages in IR, including Dutch. To address this limitation and encourage the development of Dutch IR models, we introduce BEIR-NL by automatically translating the publicly accessible BEIR datasets into Dutch. Using BEIR-NL, we evaluated a wide range of multilingual dense ranking and reranking models, as well as the lexical BM25 method. Our experiments show that BM25 remains a competitive baseline, and is only outperformed by the larger dense models trained for retrieval. When combined with reranking models, BM25 achieves performance on par with the best dense ranking models. 
In addition, we explored the impact of translation on the data by back-translating a selection of datasets to English, and observed a performance drop for both dense and lexical methods, indicating the limitations of translation for creating benchmarks. BEIR-NL is publicly available on the Hugging Face hub\footnote{\url{https://huggingface.co/collections/clips/beir-nl-6756c81a8ebab4432d922a08}}.

\end{abstract}

\section{Introduction}
An increasing number of natural language processing (NLP) tasks require an information retrieval (IR) step to identify relevant pieces of text in a large corpus of documents. Therefore, IR models are crucial in various use cases, including question-answering \cite{chen-etal-2017-reading}, claim-verification \cite{thorne2018fever}, and retrieval-augmented generation \cite{Lewis2020RetrievalAugmentedGF}. 

Recently, IR has witnessed significant progress, driven mainly by advancements in large language models (LLMs; \citealp{zhao2024dense}). Pre-trained on large corpora, these models can generate high-quality contextualized textual embeddings that capture semantic relationships beyond surface-level features like keywords. The produced vector representations demonstrate strong performance in IR tasks, as well as in other problems \cite{muennighoff2023mteb} such as classification and clustering.

Benchmarking and evaluating such models is essential in sustaining advances in NLP research.  Comprehensive benchmarks provide a standardized framework to assess the performance of models, identify their limitations, and guide the direction of future work.  BEIR (Benchmarking IR; \citealp{thakur2beir}) was introduced to address this need in IR and became a standard benchmark in zero-shot evaluation, enabling the comparison of retrieval models in a unified framework. BEIR offers a diverse and heterogeneous collection of datasets covering various domains from biomedical and financial texts to general web content, and recently has been integrated into the broader MTEB benchmark (Massive Text Embedding Benchmark; \citealp{muennighoff2023mteb}), which measures the performance of textual embeddings on a broad range of tasks. While BEIR has substantially advanced the evaluation of IR models, its main limitation lies in the monolingual structure, which restricts its application for other languages. 

In this work, we focus on extending the BEIR benchmark to Dutch, a resource-scarce language in IR research. By translating datasets from BEIR into Dutch, we aim to provide a foundation for evaluating IR models in this language. Our benchmark BEIR-NL facilitates zero-shot IR evaluation and supports the development of retrieval models tailored to Dutch. In addition, we conduct extensive evaluations of small and mid-range multilingual IR models, which support Dutch, including dense ranking and reranking models. We make the BEIR-NL benchmark available on the Hugging Face hub, ensuring that it inherits the same licenses as the datasets from BEIR (Appendix \ref{sec:appendix_a}).

\begin{table*}[htbp]
\footnotesize
\begin{center}
\begin{tabular}{llllrrr}
\toprule
\textbf{Task} & \textbf{Dataset} & \textbf{Source} & \textbf{Domain} & \textbf{\#Queries} & \textbf{\#Docs} & \textbf{Avg. D/Q} \\
\midrule
Biomedical IR & TREC-COVID & \citet{voorhees2021trec} & Biomedical & 50 & 171K & 493.5 \\
              & NFCorpus & \citet{boteva2016full} & Biomedical & 323 & 3.63K & 38.2 \\
\hline
Question Answering & NQ & \citet{kwiatkowski-etal-2019-natural} & Wikipedia & 3,452 & 2.68M & 1.2 \\
                   & HotpotQA & \citet{yang2018hotpotqa} & Wikipedia & 7,405 & 5.23M & 2.0 \\
                   & FiQA-2018 & \citet{maia201818} & Financial & 648 & 57.6K & 2.6 \\
\hline
Argument Retrieval & ArguAna & \citet{wachsmuth2018retrieval} & Miscellaneous & 1,406 & 8.67K & 1.0 \\
                   & Touche-2020 & \citet{bondarenko2020overview} & Miscellaneous & 49 & 383K & 19.0 \\
\hline
Duplicate-Question & CQADupstack & \citet{hoogeveen2015cqadupstack} & StackExchange & 13,145 & 457K & 1.4 \\
Retrieval & Quora & \citet{thakur2beir} & Quora & 10,000 & 522K & 1.6 \\
\hline
Entity Retrieval & DBPedia & \citet{hasibi2017dbpedia} & Wikipedia & 400 & 4.64M & 38.2 \\
\hline
Citation Prediction & SciDocs & \citet{cohan-etal-2020-specter} & Scientific & 1,000 & 25.7K & 4.9 \\
\hline
Fact Checking & SciFact & \citet{wadden-etal-2020-fact} & Scientific & 300 & 5.18K & 1.1 \\
              & FEVER & \citet{thorne2018fever} & Wikipedia & 6,666 & 5.42M & 1.2 \\
              & Climate-FEVER & \citet{diggelmann2020climate} & Wikipedia & 1,535 & 5.42M & 3.0 \\
\midrule
Passage Retrieval & mMARCO & \citet{bonifacio2021mmarco} & Miscellaneous & 6,980 & 8.84M & 1.1 \\
\bottomrule
\end{tabular}
\caption{Statistics of datasets included in the BEIR-NL benchmark (plus mMARCO). The table highlights the number of queries and documents, as well as the average number of relevant documents per query (Avg. D/Q) (from \citet{thakur2beir}).}
\label{tab:dataset-statistics}
\end{center}
\end{table*}

\section{Related Work}
Recently, increasing efforts have been directed towards extending English or multilingual benchmarks to cover more languages. These efforts are primarily divided into two categories: (i) the existing (or to-be) human-annotated datasets are compiled into benchmarks, or (ii) existing benchmarks are automatically translated into new languages. The first approach provides high-quality datasets but requires substantial time and financial investment. The second approach is faster and more cost-effective, but the quality of translations can affect the overall quality of the benchmark and potentially lead to inaccurate model evaluations \cite{englander2024m2qa}. However, the recent availability of relatively cheap and high-quality machine translation solutions (thanks mainly to the LLM developments and advances) has made this an attractive and commercially feasible option, especially for large datasets and benchmarks. Below we outline relevant work focused on extending existing benchmarks to additional languages.

In generative benchmarking, \citet{lai2023okapi} utilized ChatGPT to translate three widely-used benchmark datasets for LLMs into 26 languages, to evaluate the performance of models for the Okapi framework. These datasets include ARC \cite{clark2018think}, HellaSwag \cite{zellers2019hellaswag}, and MMLU \cite{hendrycksmeasuring}. \citet{vanroy2023language} extended these datasets, along with TruthfulQA \cite{lin2022truthfulqa}, to Dutch using ChatGPT. Subsequently, \citet{thellmann2024towards} added GSM8K \cite{cobbe2021training} to the mentioned benchmarking datasets and translated the entire collection into 21 European languages using DeepL.

Another branch of work focuses on extending MTEB \cite{muennighoff2023mteb}, which evaluates the quality of textual embeddings across multiple tasks. \citet{xiao2023c} extended this benchmark to Chinese (C-MTEB) by collecting  35 publicly-available Chinese datasets. MTEB-French \cite{ciancone2024mteb} added 18 datasets in French to MTEB, including both original and DeepL-translated data. Building on MTEB, \citet{wehrli2024german} introduced six benchmarking datasets for clustering text embeddings in German. A Polish version, MTEB-PL \cite{poswiata2024pl}, consists of 28 datasets, with its retrieval part sourced from BEIR-PL \cite{wojtasik2024beir}. ruMTEB \cite{snegirev2024russian} comprises 23 tasks in the MTEB format, with primarily original datasets in Russian, and with one translated using DeepL. SEB (Scandinavian Embedding Benchmark; \citealp{enevoldsen2024scandinavian}) represents 24 evaluation tasks for Scandinavian languages, incorporating a portion of existing translated datasets from MTEB.

Finally in IR, mMARCO \cite{bonifacio2021mmarco} extended the popular MSMARCO dataset \cite{bajaj2016ms} to multiple languages by translating queries and passages using Google Translate and Helsinki-NLP models \cite{tiedemann2020opus}. Most related to our work, BEIR-PL \cite{wojtasik2024beir} translated a subset of the BEIR benchmark to Polish using Google Translate. 

These efforts highlight the necessity of extending existing benchmarks to a multilingual context, enabling the evaluation of models across a wide range of languages. Building on the previous work, our study extends the BEIR benchmark to Dutch using machine translation, providing a valuable resource for evaluating IR models in this language.

\begin{table*}[ht]
\small
\begin{center}

\begin{tabular}{lccccc}
\toprule
\textbf{Model} & \textbf{Based on} & \textbf{\#Parameters} & \textbf{Dim} & \textbf{Max input} & \textbf{IR Finetuned}  \\
\midrule
e5-multilingual-small & Multilingual-MiniLM & 118M & 384 & 512 & Yes \\
e5-multilingual-base & XLMRoberta-base & 278M & 768 & 512 & Yes \\
e5-multilingual-large & XLMRoberta-large & 560M & 1024 & 512 & Yes \\
e5-multilingual-large-instruct & XLMRoberta-large & 560M & 1024 & 512 & Yes \\
gte-multilingual-base & - & 305M & 768 & 8192 & Yes \\
jina-embeddings-v3 & XLMRoberta-large & 572M & 1024 & 8192 & Yes \\
bge-m3  & XLMRoberta-large & 568M & 1024 & 8192 & Yes \\
dpr-xm & XMOD & 852M (277M\textsuperscript{\textdagger}) & 768 & 512 & Yes \\
LEALLA-small & LaBSE (distilled) & 69M & 128 & 512 & No \\
LEALLA-base & LaBSE (distilled) & 107M & 192 & 512 & No \\
LaBSE & - & 471M & 768 & 512 & No \\
mContriever & Bert-multilingual-base & 179M & 768 & 512 & No \\
\midrule
bge-reranker-v2-m3 & bge-m3 & 568M & 1024 & 8192 & Yes \\
jina-reranker-v2-base-multilingual & XLMRoberta-base & 278M & 768 & 1024 & Yes \\
gte-multilingual-reranker-base & gte-multilingual-base & 305M & 768 & 8192 & Yes \\
\bottomrule
\end{tabular}
\caption{Dense ranking (top) and reranking (bottom) models used in our experiments. `Dim' is the dimension of the output embedding vector. \texttt{LaBSE} and \texttt{gte-multilingual-base} are trained from scratch. \texttt{LEALLA} is distilled from \texttt{LaBSE}, and the rest are fine-tuned from the model mentioned in the second column. \textdagger: \texttt{dpr-xm} is modular and uses 277M parameters during inference.}
\label{tab:models}
\end{center}
\end{table*}

\section{Dataset}
The original BEIR benchmark \cite{thakur2beir} comprises 18 datasets, covering 9 different information retrieval tasks. Of these, 4 datasets are not publicly available, and therefore are removed from our selection for BEIR-NL. The remaining 14 datasets are listed in Table \ref{tab:dataset-statistics} along with their selected features and statistics. Since most retrieval models are trained on MSMARCO \cite{bajaj2016ms}, we also report on its Dutch-translated version from mMARCO \cite{bonifacio2021mmarco}, but do not include it for translation. We refer the reader to the BEIR paper \cite{thakur2beir} for further descriptions and more details on each dataset.  

\subsection{Translation}
The next step is translating the selected 14 datasets from English to Dutch. After considering commonly used options, we opted for \texttt{Gemini-1.5-flash}\footnote{A small portion of translations were done using GPT-4o-mini and Google Translate, as Gemini declined to translate certain content and had occasional issues with tags in prompts.} which offers a good balance of speed, cost, and translation quality. We prompted the model to translate the inputs, providing it with the input type (query or document), and domain (4th column in Table \ref{tab:dataset-statistics}) as context. We used the API in batch mode, which lowers the total cost to less than 450 Euro. The exact prompts can be found in Appendix \ref{sec:appendix_b}. 

To assess the translation quality, we randomly sampled 10 items from each dataset (140 in total) and asked a native Dutch speaker to check the translations against the original English text, and annotate instances for major (i.e. translation includes semantic addition or omission) or minor (i.e. translation is correct but too literal) issues. The results show  major and minor issues in 2.2\% and 14.8\% of samples respectively, which means that almost 98\% of the translated samples can be trusted for semantic accuracy. We will revisit this issue in the discussion section.      

\section{Experimental Setup}
This section provides an overview of the experimental setup used to assess the performance of different models on BEIR-NL. We mostly follow the BEIR official repository\footnote{\url{https://github.com/beir-cellar/beir}} for zero-shot evaluation, using the provided code as much as possible but occasionally adapt it to specific requirements of the evaluated models. In the following, we describe the models, data processing steps, and evaluation metrics used in our experiments.

\subsection{Models}
We include models from three categories: lexical models, dense ranking models, and dense reranking models. 

\subsubsection{Lexical models}
As the most popular lexical retrieval solution, BM25 \cite{robertson-1994-okapi} relies on keyword matching and utilizes empirical word (or token) weighting schemes to determine the relevance of documents to a given query. Despite lexical gap issues, where the vocabulary used in queries can differ from that of relevant documents, BM25 remains a robust baseline for many retrieval tasks and was outperformed only recently by E5 \cite{wang2022text} on the BEIR retrieval benchmark \cite{thakur2beir} in zero-shot setting. Similarly to \citet{wojtasik2024beir}, we utilize the BM25 implementation from Elasticsearch for Dutch.

\subsubsection{Dense ranking models}
Dense ranking (or embedding) models encode an input sequence into a dense vector, which can be used to calculate similarity or relevance between sequences (query and document in our case). Inspired by recent related studies and the MTEB leaderboard\footnote{\url{https://huggingface.co/spaces/mteb/leaderboard}}, we select the following multilingual retrieval models for our zero-shot experiments\footnote{Due to computational limitations, we exclude larger models like e5-mistral-7b-instruct and bge-multilingual-gemma2.}: mContriever \cite{izacardunsupervised}, LaBSE \cite{feng2022language}, LEALLA \cite{mao-nakagawa-2023-lealla},  mE5 \cite{wang2024multilingual}, BGE-M3 \cite{bgem3}, DPR-XM \cite{louis2024colbert}, jina-embeddings-v3 \cite{sturua2024jina}, and mGTE \cite{zhang2024mgte}. Table \ref{tab:models} lists these models along with a number of relevant features. Following the convention, we do not impose any limits on the input length for these models, allowing them to handle truncation if necessary\footnote{Considering the average document length in BEIR datasets, truncation is rarely needed for any of these models.}. In all cases, cosine similarity is employed to score similarity between the normalized embeddings. 

\subsubsection{Zero-shot reranking models}
Unlike ranking models that are employed in a bi-encoder setting, reranking models rely on cross-encoding the query and document, which can provide more accurate results at a higher computational cost. Consequently, reranking models are usually applied on the top outputs of a fast ranking model such as BM25. 

We examine three popular multilingual reranking models, namely bge-reranker-v2-m3 \cite{bgem3}, jina-reranker-v2-base-multilingual \cite{sturua2024jina}, and gte-multilingual-reranker-base \cite{zhang2024mgte} (see Table \ref{tab:models}-bottom). Following the convention \cite{thakur2beir}, we apply these models on the top-100 documents retrieved by BM25, and evaluate the reranked output. We do not restrict the input length for the reranking models, leaving them to manage truncation.

\subsection{Metrics}
To assess the performance of our models, we employ two standard retrieval metrics: nDCG@10 and Recall@100. NDCG (normalized discounted cumulative gain) is a ranking-aware metric often used to report retrieval performance, especially on graded (non-binary) labels \cite{thakur2beir}. We also report recall, which, although ranking-agnostic, is a useful and relevant metric for practical settings like retrieval-augmented generation.  

\begin{table*}[htbp]
\scriptsize
\begin{center}
\setlength{\tabcolsep}{2.8pt}
\setlength\belowcaptionskip{-20pt}
\begin{tabular}{lllcccccccccccccccc}
   \toprule
   \small{\textbf{{\scriptsize Model}}} & & &
   \rotatebox{90}{\scriptsize{MSMARCO}} &
   \rotatebox{90}{\scriptsize{TREC-COVID}} & \rotatebox{90}{\scriptsize{NFCorpus}} &
   \rotatebox{90}{\scriptsize{NQ}} & \rotatebox{90}{\scriptsize{HotpotQA}} &
   \rotatebox{90}{\scriptsize{FiQA-2018}} & \rotatebox{90}{\scriptsize{ArguAna}} & \rotatebox{90}{\scriptsize{Touche-2020}} 
   & \rotatebox{90}{\scriptsize{CQADupstack}} & \rotatebox{90}{\scriptsize{Quora}} 
   & \rotatebox{90}{\scriptsize{DBPedia}} &
   \rotatebox{90}{\scriptsize{SciDocs}} & \rotatebox{90}{\scriptsize{SciFact}} &
   \rotatebox{90}{\scriptsize{FEVER}} & \rotatebox{90}{\scriptsize{Climate-FEVER}} \\  
   \midrule
   {\scriptsize BM25} & \multirow{16}{*}{\rotatebox{90}{{\scriptsize NDCG@10}}} &   &  16.87 & 63.37 &  30.54 & 25.09 & 53.62 & 18.73 & 41.76 & 28.15 & 27.77 & 65.92 & 25.46 & 11.44 & 61.13 & 60.65 & 12.09\\[3pt]
   {\scriptsize multilingual-e5-small}  & &    & 30.85\textsuperscript{\textdagger} & 41.74 & 24.10 & 27.03\textsuperscript{\textdagger} & 53.30\textsuperscript{\textdagger} & 20.39 & 44.76 & 16.04 & 28.51 & 79.85\textsuperscript{\textdagger} & 25.89 & 6.58 & 58.82 & 56.69\textsuperscript{\textdagger} & 14.08 \\[3pt]
   {\scriptsize multilingual-e5-base}  & &    & 32.79\textsuperscript{\textdagger} & 40.68 & 24.17 & 36.06\textsuperscript{\textdagger} & 60.87\textsuperscript{\textdagger} & 23.76 & 47.06 & 10.29 & 30.36 & 81.02\textsuperscript{\textdagger} & 28.74 & 10.53 & 67.23 & 58.52\textsuperscript{\textdagger} & 16.31\\[3pt]
   {\scriptsize multilingual-e5-large}  & &    & \textbf{37.51}\textsuperscript{\textdagger} & 69.72 & 28.06 & 49.15\textsuperscript{\textdagger} & 67.95\textsuperscript{\textdagger} & 31.84 & 48.90 & 22.18 & 31.92 & 82.01\textsuperscript{\textdagger} & \textbf{38.67} & 11.95 & 68.38 & 72.73\textsuperscript{\textdagger} &13.76 \\[3pt]
   {\scriptsize multilingual-e5-large-instruct}  & &    & 34.35\textsuperscript{\textdagger} & 71.22 & 31.08 & \textbf{55.79}\textsuperscript{\textdagger} & 65.97\textsuperscript{\textdagger} & \textbf{37.93} & 50.32 & 26.67 & 36.95 & 83.54\textsuperscript{\textdagger} & 38.24 & \textbf{18.07} & 69.10 & 79.39\textsuperscript{\textdagger} & 21.05 \\[3pt]
   {\scriptsize gte-multilingual-base}      & &    & 27.19\textsuperscript{\textdagger} & 53.36 & 27.97\textsuperscript{\textdagger} & 47.42\textsuperscript{\textdagger} & 58.53\textsuperscript{\textdagger} & 29.45 & \textbf{52.85}\textsuperscript{\textdagger} & 22.60 \textsuperscript{\textdagger} & 31.59\textsuperscript{\textdagger} & 81.25\textsuperscript{\textdagger} & 36.46\textsuperscript{\textdagger} & 15.86 & 64.41 & 82.68\textsuperscript{\textdagger} & 17.53\\[3pt]
   {\scriptsize jina-embeddings-v3}       & &    & 26.05\textsuperscript{\textdagger} & 54.46 & 29.84 & 37.26\textsuperscript{\textdagger} & 51.82 & 35.71 & 52.23 & 15.05 & 36.16 & 82.92 & 30.71 & 18.42 & 64.90 & 68.88 & 19.54 \\[3pt]
    {\scriptsize bge-m3}       & &    & 31.96\textsuperscript{\textdagger} & 48.22 & 27.90 & 51.92\textsuperscript{\textdagger} & 65.20\textsuperscript{\textdagger} & 32.60 & 52.16 & 22.68 &  34.75 & \textbf{83.72} & 35.46 & 14.41 & 62.83 & 76.08 & \textbf{26.39} \\[3pt]   
    {\scriptsize dpr-xm}       & &    & 28.46\textsuperscript{\textdagger} & 40.86 & 18.58 & 28.56 & 26.34 & 13.98 & 26.91 & 15.99 & 18.73 & 74.70 & 21.07 & 8.64 & 34.29 & 49.46 & 11.16\\[3pt]
   {\scriptsize LEALLA-small} & &    & 3.95 & 13.32 & 5.56 & 5.11 & 12.18 & 3.41 & 19.25 & 5.65 & 13.14 & 68.50 & 9.60 & 3.70 & 12.98 & 7.08 & 0.34\\[3pt]
   {\scriptsize LEALLA-base}     & &    & 5.60 & 14.44 & 6.09 & 7.77 & 17.46 & 3.75 & 24.97 & 5.00 & 14.34 & 70.87 & 13.40 & 3.09 & 7.13 & 7.46 & 1.15\\[3pt]    
    {\scriptsize LaBSE}       & &    & 6.87 & 18.50 & 13.54 & 11.24 & 18.64 & 7.38 & 39.15 & 4.67 & 19.66 & 75.55 & 15.27 & 6.32 & 39.07 & 12.51 & 3.85\\[3pt]
    {\scriptsize mContriever}       & &    & 7.46\textsuperscript{\textdagger} & 17.51 & 13.36 & 10.50 & 27.84 & 5.41 & 39.60 & 6.15 & 12.81 & 72.90 & 15.58 & 4.93 & 37.89 & 21.51 & 3.08 \\[3pt]     
    {\scriptsize BM25 + bge-reranker}       & &    &  31.80\textsuperscript{\textdagger} & 76.47 & \textbf{33.78} & 51.28\textsuperscript{\textdagger} & \textbf{71.78}\textsuperscript{\textdagger} & 30.41 & 47.27 &  \textbf{33.78} & 31.70 &  76.81 & 37.84 & 13.88 &  69.94 & 84.17 & 25.60\\[3pt]
    {\scriptsize BM25 + jina-reranker}       & &    & 31.93\textsuperscript{\textdagger} &  \textbf{76.83} & 33.19 & 49.07\textsuperscript{\textdagger} & 70.57 &  30.86 &  48.53 & 30.96 & 34.06 &  79.44 &  36.26 & 14.49 &  \textbf{70.68} & \textbf{85.17} & 22.56 \\[3pt]
    {\scriptsize BM25 + gte-reranker}       & &    & 28.90\textsuperscript{\textdagger} & 76.24 & 28.26\textsuperscript{\textdagger} & 47.85\textsuperscript{\textdagger} & 70.43\textsuperscript{\textdagger} & 24.13 & 46.74\textsuperscript{\textdagger} & 28.26\textsuperscript{\textdagger} & 25.69\textsuperscript{\textdagger} & 74.95\textsuperscript{\textdagger} & 36.67\textsuperscript{\textdagger} & 13.22 & 68.37 &  85.13\textsuperscript{\textdagger} & 22.96\\[3pt]
   \midrule
{\scriptsize BM25} & \multirow{16}{*}{\rotatebox{90}{{\scriptsize Recall@100}}} &   &  51.20 & 10.52 & 22.16 & 65.57 & 70.54 & 42.83 & 92.32 & 44.16 & 54.77 & 88.66 & 36.92 & 26.49 & 83.42 & 89.20 & 30.42 \\[3pt]
   {\scriptsize multilingual-e5-small}  & &    & 74.63\textsuperscript{\textdagger} & 7.89 & 23.56 & 60.70\textsuperscript{\textdagger} & 69.45\textsuperscript{\textdagger} & 47.10 & 94.59 & 38.18 & 56.99 & 97.51\textsuperscript{\textdagger} & 35.83 & 22.93 & 87.67 & 85.83\textsuperscript{\textdagger} & 40.47 \\[3pt]
   {\scriptsize multilingual-e5-base}  & &    & 77.39\textsuperscript{\textdagger} & 6.58 & 22.09 & 73.61\textsuperscript{\textdagger} & 76.24\textsuperscript{\textdagger} & 55.02 & 95.59 & 32.96 & 60.65 & 97.93\textsuperscript{\textdagger} & 39.40 & 29.78 & 91.00 & 89.98\textsuperscript{\textdagger} & 42.69\\[3pt]
   {\scriptsize multilingual-e5-large}  & &    & \textbf{82.71}\textsuperscript{\textdagger} & 13.31 & 27.34 & 83.49\textsuperscript{\textdagger} & \textbf{82.21}\textsuperscript{\textdagger} & 61.81 & 96.37 & 43.65 & 63.30 & 98.66\textsuperscript{\textdagger} & 47.26 & 30.42 & 92.27 & 93.08\textsuperscript{\textdagger} & 32.68 \\[3pt]
   {\scriptsize multilingual-e5-large-instruct}  & &    & 80.89\textsuperscript{\textdagger} & \textbf{14.48} &  \textbf{28.88} & \textbf{92.39}\textsuperscript{\textdagger} & 80.55\textsuperscript{\textdagger} & 68.70 & 98.86 & 46.97 & 70.56 & 98.83\textsuperscript{\textdagger} & \textbf{49.66} & 40.80 & \textbf{93.67} & \textbf{94.53}\textsuperscript{\textdagger} & \textbf{46.05}\\[3pt]
   {\scriptsize gte-multilingual-base}      & &    & 70.29\textsuperscript{\textdagger} & 10.74 & 27.89\textsuperscript{\textdagger} & 85.39\textsuperscript{\textdagger} & 70.08\textsuperscript{\textdagger} & 61.53 & 97.87\textsuperscript{\textdagger} & 41.12\textsuperscript{\textdagger} & 66.14\textsuperscript{\textdagger} & 98.12\textsuperscript{\textdagger} & 44.11\textsuperscript{\textdagger} & 37.43 & 91.00 & 94.32\textsuperscript{\textdagger} & 40.40\\[3pt]
   {\scriptsize jina-embeddings-v3}       & &    & 73.43\textsuperscript{\textdagger} & 11.74 & 26.50 & 84.43\textsuperscript{\textdagger} & 68.04 & \textbf{69.98} & \textbf{98.93} & 37.69 & \textbf{72.62} & 98.58 & 42.22 & \textbf{42.64} & 91.17 & 93.04 & 44.98 \\[3pt]
    {\scriptsize bge-m3}       & &    & 77.71\textsuperscript{\textdagger} & 9.43 & 25.20 & 89.62\textsuperscript{\textdagger} & 80.20\textsuperscript{\textdagger} & 63.41 & 97.44 & \textbf{48.70} & 66.89 & \textbf{98.85} & 46.30 & 35.02 & 91.93 & 94.11 & 56.54\\[3pt] 
    {\scriptsize dpr-xm}       & &    & 67.77\textsuperscript{\textdagger} & 5.78 & 17.95 & 62.42 & 38.31 & 33.81 & 78.73 & 36.46 & 41.94 & 93.41 & 22.25 & 19.36 & 67.26 & 76.17 & 28.54\\[3pt]
    {\scriptsize LEALLA-small} & &    & 15.99 & 1.44 & 9.12 & 19.99 & 23.48 & 12.19 & 56.47 & 9.89 & 32.58 & 91.41 & 13.38 & 12.62 & 42.81 & 14.79 & 1.81\\[3pt]
   {\scriptsize LEALLA-base}     & &    & 22.12 & 1.61 & 9.92 & 27.45 & 30.32 & 13.04 & 61.30 & 8.39 & 33.89 & 93.13 & 18.80 & 10.73 & 34.18 & 14.97 & 2.61\\[3pt]
    {\scriptsize LaBSE}       & &    & 26.71 & 1.97 & 16.05 & 41.68 & 33.56 & 25.57 & 87.98 & 10.09 & 47.06 & 95.87 & 22.91 & 21.50 & 74.67 & 36.48 & 15.24 \\[3pt]
    {\scriptsize mContriever}       & &    & 32.06\textsuperscript{\textdagger} & 1.71 & 16.81 & 40.42 & 45.97 & 20.36 & 91.61 & 12.06 & 35.91 & 94.48 & 25.25 & 18.56 & 74.24 & 48.31 & 10.29 \\[3pt]       
   \bottomrule
\end{tabular}
\caption{Performance of selected models on the BEIR-NL benchmark (plus MSMARCO), measured by NDCG@10 (top) and Recall@100 (bottom).{\textdagger} indicates results that are (or are highly likely to be) inflated because of potential contamination of the model with in-domain data for a given dataset, based on available descriptions from the corresponding work (i.e. they are highly unlikely to be zero-shot). \texttt{bge-reranker}, \texttt{jina-reranker}, and \texttt{gte-reranker} refer to \texttt{bge-reranker-v2-m3}, \texttt{jina-reranker-v2-base-multilingual}, and \texttt{gte-multilingual-reranker} models, respectively.
}
\label{tab:general}
\end{center}
\end{table*}

\section{Results and Discussion}
\subsection{Retrieval Performance on BEIR-NL}
Table \ref{tab:general} shows the retrieval performance of the selected models on the 14 subsets of BEIR-NL, in addition to MSMARCO. As mentioned before, MSMARCO is not part of our dataset, but considering its popularity in retrieval training, we include it in the evaluations (based on the Dutch-translated version from mMARCO \cite{bonifacio2021mmarco}). 

The results show that BM25 still provides a competitive baseline, and in many cases is only outperformed by the larger dense models. The four recently released multilingual-e5-large-instruct, gte-multilingual-base, jina-embeddings-v3 and bge-m3 achieve the best overall performances, with multilingual-e5-large-instruct getting the highest Recall@100 on half of the datasets. We also observe a sizeable gap between the older `sentence embedding' models, and the new generation of trained-for-retrieval models (see the last column in Table \ref{tab:models}), with the latter achieving substantially higher results. However, based on their published metadata, the majority of these models have been at least partially exposed to BEIR datasets in their training process, which makes the comparison unfair (The corresponding \textit{potentially inflated} results are marked with a {\textdagger} in the table.). In other words, in these cases the evaluation could not be considered proper zero-shot.

Finally, the last three rows of the top section in Table \ref{tab:general} (NDCG@10 results) show the performance of the reranking models when used in combination with BM25 as the first-step ranker. As demonstrated, this approach can often offer a competitive edge over the best ranking models.

\begin{table*}[tbp]
\small 
\begin{center}
\begin{tabular}{llccccccccccc}
\toprule
\textbf{Metric} & \textbf{Benchmark} & \rotatebox{90}{\scriptsize TREC-COVID} & 
\rotatebox{90}{\scriptsize NFCorpus} & \rotatebox{90}{\scriptsize NQ} & 
\rotatebox{90}{\scriptsize HotpotQA} & \rotatebox{90}{\scriptsize FiQA-2018} & 
\rotatebox{90}{\scriptsize ArguAna} & \rotatebox{90}{\scriptsize CQADupstack} & 
\rotatebox{90}{\scriptsize DBPedia} & \rotatebox{90}{\scriptsize SciDocs} & 
\rotatebox{90}{\scriptsize SciFact}  & \textbf{Average} \\
\midrule
NDCG@10 & \texttt{BEIR-NL}   & 63.4 & 30.5 & 25.1 & 53.6 & 18.7 & 41.8 & 27.8 & 25.6 & 11.4 & 61.1 & 35.9\\
        & \texttt{BEIR-PL}   & 61.0 & 31.9 & 20.1 & 49.2 & 19.0 & 41.4 & 28.4 & 22.9 & 14.1 & 62.5 & 35.1\\
        & \texttt{BEIR (EN)} & 68.9 & 34.3 & 32.6 & 60.2 & 25.4 & 47.2 & 32.5 & 32.1 & 16.5 & 69.1 & 41.9\\
\midrule
Recall@100 & \texttt{BEIR-NL} & 10.5 & 22.2 & 65.6 & 70.5 & 42.8 & 92.3 & 54.8 & 36.9 & 26.5 & 83.4 & 50.6\\
           & \texttt{BEIR-PL} & 10.1 & 24.6 & 57.9 & 67.1 & 44.1 & 93.5 & 53.9 & 30.1 & 33.0 & 88.4 & 50.3\\
           & \texttt{BEIR (EN)} & 11.7 & 26.0 & 78.3 & 76.3 & 54.9 & 95.2 & 62.1 & 43.5 & 36.8 & 92.0 & 57.7\\
\bottomrule
\end{tabular}
\end{center}
\caption{BM25 performance on the overlapping subset of BEIR-NL, BEIR-PL, and original BEIR, for which performance data is publicly available. Results for BEIR-PL and BEIR are  from \citet{wojtasik2024beir}.}
\label{tabel:bm25_multilingual}
\end{table*}

\begin{table*}[tbp]
\small 
\begin{center}
\begin{tabular}{llcccccccccc}
\toprule
\textbf{Metric} & \textbf{Benchmark} &
\rotatebox{90}{\scriptsize{TREC-COVID}} & \rotatebox{90}{\scriptsize{NFCorpus}} &
\rotatebox{90}{\scriptsize{NQ}} & \rotatebox{90}{\scriptsize{HotpotQA}} &
\rotatebox{90}{\scriptsize{FiQA-2018}} & \rotatebox{90}{\scriptsize{ArguAna}} &
\rotatebox{90}{\scriptsize{DBPedia}} & \rotatebox{90}{\scriptsize{SciDocs}} &
\rotatebox{90}{\scriptsize{SciFact}} & \textbf{Average} \\
\midrule
NDCG@10 & \texttt{BEIR-NL} & 53.4 & 28.0 & 47.4 & 58.5 & 29.4 & 52.9 & 36.5 & 15.9 & 64.4 & 42.9\\
        & \texttt{BEIR-PL} & 59.4 & 26.8 & 43.1 & 56.9 & 29.0 & 53.2 & 32.5 & 14.2 & 58.9 & 41.6 \\
        & \texttt{BEIR (EN)} & 57.6 & 36.6 & 58.1 & 63.0 & 45.0 & 58.2 & 40.1 & 18.2 & 73.4 & 50.0\\
\bottomrule
\end{tabular}
\end{center}
\caption{Performance of \texttt{gte-multilingual-base} on the overlapping subset of BEIR-NL, BEIR-PL, and original BEIR, for which performance data is publicly available. Results for BEIR-PL and BEIR are sourced from the MTEB leaderboard.}
\label{tabel:gte_multilingual}
\end{table*}

\subsection{Comparison with BEIR and BEIR-PL}
Since BEIR-NL is a translated benchmark, we can compare the performance of the retrieval methods on parallel subsets in different languages, including the (translated) Polish version, BEIR-PL \cite{wojtasik2024beir}. 

Tables \ref{tabel:bm25_multilingual} and \ref{tabel:gte_multilingual} show this comparison for BM25 and gte-multilingual-base, across the subsets for which performance data is publicly available\footnote{BEIR-PL only covers 10 of the 14 public BEIR datasets.}.
As Table \ref{tabel:bm25_multilingual} reveals, BM25 performs comparably on BEIR-NL and BEIR-PL subsets, with a marginal overall advantage for BEIR-NL. However, these numbers lag behind the BM25 performance on the original BEIR dataset by 6-7 points in NDCG@10 and Recall@100. One potential reason for this drop is the lexical mismatch between
the translated query and relevant passages since queries and passages are translated independently\footnote{Assuming a uniform BM25 performance for different languages, which is not trivial.} \cite{bonifacio2021mmarco}. Table \ref{tabel:gte_multilingual} shows that the performance difference persists with dense models (e.g. gte-multilingual-base). Here, the discrepancy can be attributed to both the data (translation quality) and model (higher competence in English compared to other languages).

\subsection{Impact of Translation}
To isolate the semantic effect of translation (from that of the model/language) we back-translate a subset of 5 BEIR-NL datasets to English using the same translation pipeline, and compare the performance of lexical and dense models on this version against the original one. Table \ref{tabel:bm25_selected} shows the results (NDCG@10), which indicate an average drop of 1.9 and 2.6 points for the lexical (BM25) and dense model (gte-multilingual-base) respectively. Since the model-language competence factor is absent here, this drop can be considered a proxy for the impact of translation on the benchmark quality and/or reliability.

\begin{table*}[tbp]
\small
\begin{center}
\begin{tabular}{llccccccc}
\toprule
   \textbf{Model} & \textbf{BEIR}  & \rotatebox{90}{\scriptsize{NFCorpus}} &
   \rotatebox{90}{\scriptsize{FiQA-2018}} & \rotatebox{90}{\scriptsize{ArguAna}} &
   \rotatebox{90}{\scriptsize{SciDocs}} & \rotatebox{90}{\scriptsize{SciFact}}  & \textbf{Average} & {\textbf{{ $\Delta_{tr}$}}}\\
   \midrule
    BM25 & original & 34.3 & 25.4 & 47.2 & 16.5 & 69.1 & 38.5 & -\\
     & back-translated & 32.4 & 22.0 & 45.2 & 15.1 & 68.2 & 36.6 & -1.9\\

   \midrule
    gte-multilingual-base & original & 36.7 & 45.0 & 58.2 & 18.2 & 73.4 & 46.3 & -\\ 
     & back-translated & 32.6 & 40.7 & 55.0 & 18.3 & 71.7 & 43.7& -2.6 \\
    
   \bottomrule
\end{tabular}
\end{center}
\caption{NDCG@10 results for \texttt{BM25} and \texttt{gte-multilingual-base} on selected datasets from the original BEIR, and their back-translated version (from Dutch to English). $\Delta_{tr}$ is the change in average performance due to back translation. }
\label{tabel:bm25_selected}
\end{table*}

\section{Conclusions and Future Work}
In this work, we introduced BEIR-NL, an automatically translated version of the BEIR benchmark into Dutch, which aims to address the need for the evaluation of IR models in this language. Using BEIR-NL, we conducted extensive zero-shot evaluations for various models, including one lexical model as well as small and mid-range dense retrieval and reranking models. These experiments showed that larger dense IR models generally outperform BM25, while BM25 remains a competitive baseline for smaller models. Furthermore, combining BM25 with reranking models results in performance comparable to the best dense retrieval models.

We also observed several challenges, including the impact of translation on retrieval performance and the risk of in-domain data contamination in IR models. These issues might affect the reliability of zero-shot evaluations on this benchmark
and highlight the need for creating native Dutch resources, which we leave for future work. 

BEIR-NL fills a critical gap in the evaluation of Dutch IR models and sets a foundation for further development of IR benchmarks in Dutch. By making BEIR-NL publicly available, we aim to support future research and encourage the development of retrieval models for this language.

\subsection*{Limitations}
Besides the issues originated from translation (which we briefly addressed before), here we discuss other important limitations pertinent to this work.\\ 
\textbf{Native Dutch Resources.} While BEIR-NL provides a benchmark for evaluating IR models in Dutch, it relies on translations from the original BEIR, which is exclusively in English. This lack of native Dutch datasets limits the ability of BEIR-NL to fully represent and reflect the linguistic nuances and cultural context of the language, and therefore the complexities of Dutch IR, especially in domain-specific contexts with local terminology and knowledge.\\
\textbf{Data Contamination.} Many modern IR models are trained on massive corpora that might include content from BEIR. Table \ref{tab:general} indicates multiple models that have (or might have) been exposed to in-domain contamination for a given dataset. This can result in inflated performance --as models might have already seen the relevant data during different phases of training-- raising concerns about the validity of zero-shot evaluations. Ensuring a truly zero-shot evaluation is a difficult challenge, as many IR models lack transparency regarding the exact composition of training corpora.  \\
\textbf{Benchmark Validity Over Time.} BEIR has become a standard benchmark to evaluate the performance of IR models, attracting a large number of evaluations over time. This extensive usage introduces the risk of overfitting, as researchers might unintentionally train models tailored to perform well on BEIR rather than on broader IR tasks. In addition, advances in IR models and evaluation needs might outpace the benchmark, making it less representative and less relevant. As a result, the relevance and validity of BEIR as well as BEIR-NL may diminish over time.\\
\section*{Acknowledgments}
This research received funding from the Flemish Government under the “Onderzoeksprogramma Artificiële Intelligentie (AI) Vlaanderen” programme.
We would like to thank Jens Van Nooten for assessing the quality of the translations. In addition, we acknowledge the use of the GPT-4o model for assisting with error checking and proofreading of this paper.

\bibliography{custom}

\appendix
\section{Appendix: Licenses}
\label{sec:appendix_a}

The BEIR repository on Hugging Face\footnote{\url{https://huggingface.co/datasets/BeIR/}} reports that the following datasets are distributed under the CC BY-SA 4.0 license: NFCorpus, FiQA-2018, Quora, Climate-Fever,  FEVER, NQ, DBPedia, ArguAna, Touché-2020, SciFact, SCIDOCS, HotpotQA, TREC-COVID. The only one exception is CQADupStack\footnote{\url{https://github.com/D1Doris/CQADupStack}} with the Apache License 2.0 license.

\section{Appendix: Translation Prompts }
\label{sec:appendix_b}
We prompt  \texttt{Gemini-1.5-flash} with the following instructions (temperature = 0).

\textbf{Query Prompt:}\texttt{"Translate to English the QUERY from the \{domain\} domain. Provide only the translation.
QUERY:\textbackslash n ['\{query\}']".}

\textbf{Document Prompt:}\texttt{"Translate to English the DOCUMENT from the \{domain\} domain. Provide only the translation.
DOCUMENT:\textbackslash n ['<title> \{title\} <title\textbackslash> 
<body> \{document\} <body\textbackslash>']".}

\end{document}